\documentclass[letterpaper, 10 pt, conference]{ieeeconf}  

\IEEEoverridecommandlockouts                              

\usepackage{graphicx}
\usepackage{subfigure}
\usepackage{threeparttable}

\usepackage{amsmath} 
\usepackage{amssymb}  
\usepackage{bm} %
\usepackage{verbatim}
\usepackage{epstopdf}
\usepackage{tabularx}
\usepackage{cite}
\usepackage{booktabs}
\usepackage{multirow}
\usepackage{color}

\usepackage[utf8]{inputenc}

\title{\LARGE \bf
Improving the generalization of network based relative pose regression: dimension reduction as a regularizer
}

\author{Xiaqing Ding$^{1}$, Yue Wang$^{1}$, Li Tang$^{1}$ Yanmei Jiao$^{1}$ and Rong Xiong$^{1}$
\thanks{$^{1}$Xiaqing Ding, Yue Wang, Li Tang, Yanmei Jiao, and Rong Xiong are with the State Key Laboratory of Industrial Control and Technology, Zhejiang University, Hangzhou, P.R. China. Yue Wang is the corresponding author {\tt\small wangyue@iipc.zju.edu.cn}. Rong Xiong is the co-corresponding author {\tt\small rxiong@zju.edu.cn}.}%
}

\begin{document}

\maketitle
\thispagestyle{empty}
\pagestyle{empty}

\begin{abstract}
Visual localization occupies an important position in many areas such as Augmented Reality, robotics and 3D reconstruction. The state-of-the-art visual localization methods perform pose estimation using geometry based solver within the RANSAC framework. However, these methods require accurate pixel-level matching at high image resolution, which is hard to satisfy under significant changes from appearance, dynamics or perspective of view. End-to-end learning based regression networks provide a solution to circumvent the requirement for precise pixel-level correspondences, but demonstrate poor performance towards cross-scene generalization. In this paper, we explicitly add a learnable matching layer within the network to isolate the pose regression solver from the absolute image feature values, and apply dimension regularization on both the correlation feature channel and the image scale to further improve performance towards generalization and large viewpoint change. We implement this dimension regularization strategy within a two-layer pyramid based framework to regress the localization results from coarse to fine. In addition, the depth information is fused for absolute translational scale recovery. Through experiments on real world RGBD datasets we validate the effectiveness of our design in terms of improving both generalization performance and robustness towards viewpoint change, and also show the potential of regression based visual localization networks towards challenging occasions that are difficult for geometry based visual localization methods. 
\end{abstract}

\section{Introduction}
Visual localization provides accurate orientation and position information for many applications such as Augmented Reality\cite{Klein2007Parallel, middelberg2014scalable,qin2018vins} and mobile robots \cite{tang2019topological,sattler2018benchmarking}. The state-of-the-art visual localization systems usually contain four sequential modules: image retrieval \cite{arandjelovic2016netvlad, torii201524}, feature extraction and description \cite{rublee2011orb, detone2018superpoint }, feature matching \cite{sarlin2020superglue} and pose estimation \cite{lepetit2009epnp, jiao20202}. Each module in these modular pipeline based localization ({MPL}) methods has been investigated for many years varying from traditional to learning based solutions. Concluding from the current researches in the community, in many cases learning based methods show better performance in the first three modules, while for pose estimation traditional geometry based methods  under the RANSAC frameworks\cite{lepetit2009epnp, jiao20202, sarlin2019coarse}  still hold the superiority in terms of generalization and precision. 

To localize in environments with low appearance change and sufficient texture, geometry based MPL could demonstrate superior localization performance. However, if the appearance changes significantly, or a majority of the view are textureless, those methods are prone to failure due to the inadequate matching inliers. Many researches devote to improving the precision and robustness of detecting pixel-level correspondences between images\cite{truong2020glu, sarlin2020superglue}.
But usually in these situations only coarse correspondence can be determined and is hard to find precise matches at high resolution level even with human annotation.

However, is accurate pixel-to-pixel image correspondence really necessary for pose estimation? End-to-end learning based visual localization methods propose a promising solution that could bypass it. Considering the coordinate of estimated pose, end-to-end localization could be categorized as absolute pose regression (APR) and relative pose regression (RPR) methods. APR methods directly regress the global pose of the query image \cite{kendall2015posenet, kendall2017geometric} or global 3D points for pixels on the query image \cite{brachmann2017dsac, brachmann2018learning, brachmann2019expert, li2020hierarchical}. Though some of these methods could achieve higher localization accuracy than geometry based MPL solutions \cite{brachmann2017dsac,brachmann2018learning, brachmann2019expert, li2020hierarchical}, APR methods can not generalize to unseen scenes as the scene-specific information is encoded within the models. 

In contrast to regress the variables in the global coordinate directly, relative pose regression (RPR) based methods regress the relative pose between two images \cite{laskar2017camera, balntas2018relocnet, ding2019camnet, zhou2020learn}, which can be considered as the combination of last three modules in MPL. Thus combined with image retrieval, RPR methods can achieve global localization with no need to encode the scene-specific geographic information within the network, thus possessing the potential of generalization. 

Unfortunately, currently many RPR based methods show poor generalization performance to unseen scenes as shown in experimental results \cite{zhou2020learn}. Compared with the MPL pipeline, matching and pose estimation processes are coupled in many RPR networks that regress the relative pose directly from the concatenation of the input image feature pair \cite{laskar2017camera, balntas2018relocnet, ding2019camnet}, which makes the regression results related to the enormous and scene-specific feature space. \cite{zhou2020learn} can be considered as explicitly including the matching process within the RPR pipeline by adding a learnable Neighborhood Consensus (NC) matching layer \cite{rocco2018neighbourhood} before regressing the pose. The matching layer outputs a score map that contains the entire pairwise feature correlation score between the two input images. In this way the regression result is related to the correlation between image features, which isolates the regression layer from the absolute feature values that vary across scenes. However, the generalization performance is still not acceptable, thus they infer that the implicit feature matching cannot be correctly learned within the RPR network \cite{zhou2020learn}.

\begin{figure*}[]
	\begin{center}
		\includegraphics[width=0.9\textwidth]{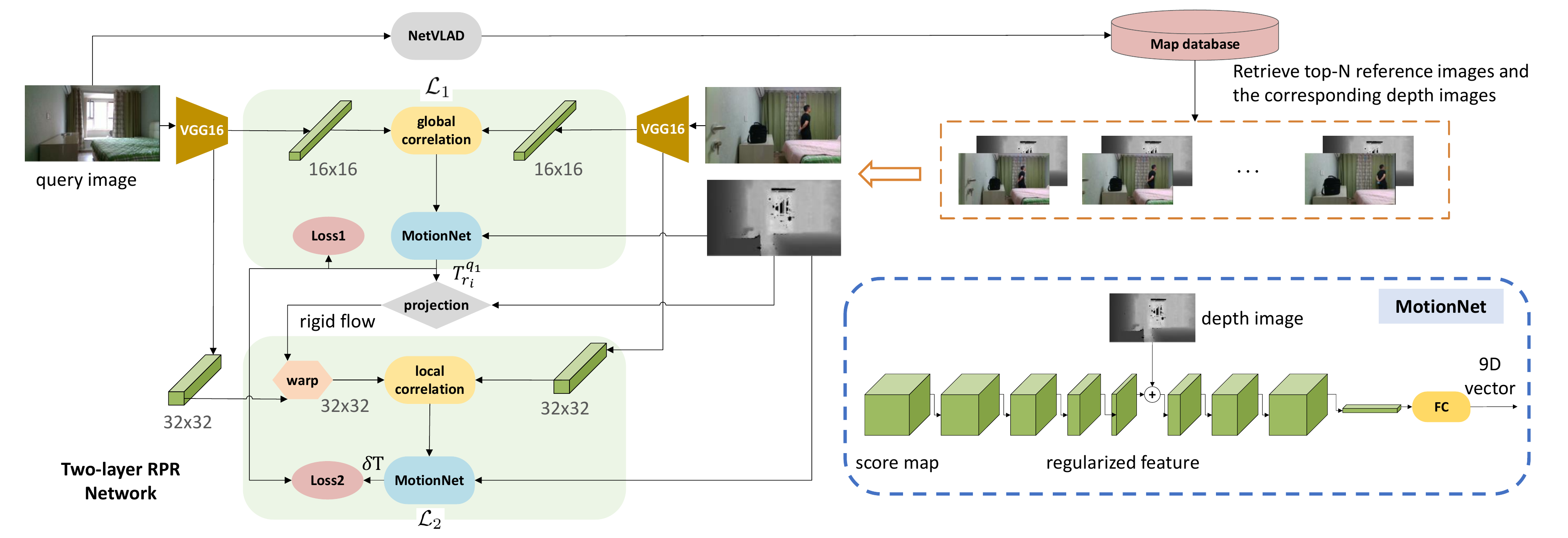}
		\caption{This figure demonstrates the whole pipeline of our visual localization framework. The left part of the figure shows our proposed two-layer relative pose regression network. We draw the detail about the regularization based pose regression layer (MotionNet) within the blue dotted box on the right.}
		\label{framework}
	\end{center}
\end{figure*}

In this paper, we argue that the implicit feature matching can be handled within the RPR network to boost generalization, and propose a novel framework to improve the performance of RPR methods. 
As pose information is the only supervision during the training process, it's difficult to apply sufficient constraint for network to learn both matching and pose regression facing huge input data dimensions. We perform regularization to reduce the dimensions of both the image scale and feature correlations aiming to reasonably apply additional constraints on the network. To do this, besides adding a matching layer \cite{rocco2018neighbourhood} to explicitly calculate correlation information, we add a convolutional neural network (CNN) with  bottleneck structure to regularize the feature correlations, and implement this new structure within a two-layer pyramid based framework to regress the relative pose from coarse to fine at low resolution with large receptive field, which further reduces the input dimension for regression.
 Moreover, depth image is concatenated with the regularized feature correlation to recover the absolute scale of the regressed pose as shown in Fig. \ref{framework}.
We implement RPR networks with different regression structures within this two-layer framework and compare their performance on public indoor RGBD datasets. Through experiments we validate the effectiveness of both the implicit matching layer and dimension regularization in terms of generalization improvement as well as the depth fusion in terms of scale recovery. Besides, the structure with correlation feature regularization shows superior performance in occasions with large perspective of view. 
The experimental results also demonstrate that in challenging changing environments learning based methods possess more potential than state-of-the-art MPL methods with geometry solvers.

\section{Related Work}
In this section we review the works of visual localization that are related to geometry and learning based solutions. For a more complete review of this area, we recommend the survey \cite{chen2020survey}.

\subsection{Geometry based visual localization}
Geometry based visual localization usually solves the image pose given matches between 2D keypoints and 3D map points using the RANSAC \cite{fischler1981random} based Perspective-n-Point (PnP) solver \cite{gao2003complete,lepetit2009epnp,jiao20202}. The matches can be computed following nearest neighbor searching according to the distance between feature descriptors on the query image and the image retrieved from the map. Recently, there are also some learning based methods solving out the matches using CNN \cite{rocco2018neighbourhood,rocco2020efficient} or Graph Neural Network (GNN) \cite{sarlin2020superglue}.   

\subsection{Learning for absolute pose estimation}
Learning based methods for absolute pose estimation encode the environmental information within the network parameters during the training process. Given a query image, some of these works directly output the global poses w.r.t the map. PoseNet \cite{kendall2015posenet} is the first end-to-end network modified from GoogleNet \cite{szegedy2015going} to regress the translation and rotation represented by quaternion of the input image, and many following works \cite{kendall2017geometric, naseer2017deep, walch2017image} are designed to improve the performance.

Regressing the pose directly end-to-end is efficient but the precision is less accurate. Scene coordinate regression based visual localization \cite{brachmann2017dsac, li2020hierarchical} chooses another way for pose estimation. Instead of directly regressing the image pose, these methods regress the global 3D location for each pixel on the query image. \cite{brachmann2017dsac} utilizes a CNN network for scene coordinate regression, and the achieved 3D-2D matches are forwarded into a novel differentiable RANSAC to achieve end-to-end training. This method could exceed the traditional geometry based methods on localization accuracy, but only shows the efficiency in small environment. There are many following methods designed to improve its performance by adding reprojection \cite{brachmann2018learning} and multi-view geometry constrains \cite{cai2019camera}. To extend methods to large and ambiguous  environments, some methods leverage hierarchical network structure to regress the 3D scene coordinates from coarse to fine \cite{li2020hierarchical}, or integrate DSAC within a Mixture of Expert (MoE) framework \cite{brachmann2019expert}. However, as the map information is encoded within the parameters, these methods can not be generalized to unseen scenes.   

\subsection{Learning for relative pose estimation}
Learning the relative pose between two images is a more general solution to achieve global localization. The reference map images are retrieved by pre-trained networks \cite{zhou2020learn} or networks that are jointly trained with the following RPR parts \cite{balntas2018relocnet, ding2019camnet}. RPR problem is also studied in some visual odometry works \cite{yin2018geonet, wang2020flow}, but in the context of localization the photometric consistency is usually broken. 

In many RPR networks the depth information is not utilized for localization, thus the estimated translation is up-to-scale, and absolute localization results have to be recovered by RANSAC based triangulation \cite{zhou2020learn,laskar2017camera}. In this paper we fuse the depth information within regression and show its ability to recover pose with scale.

\section{Method}
Our goal is to achieve robust visual localization given the current query image $I_q\in{\mathbb{R}^{H\times W\times 3}}$ and the map $\mathcal{M}$ constructed with RGB images $\{I_r\in{\mathbb{R}^{H\times W\times 3}}\}$ with corresponding depth $\{D_r\in\mathbb{R}^{H\times W\times1}\}$. To achieve this, a two-stage visual localization pipeline is utilized to first retrieve top-$N$ ranked images from the map, then estimate the relative poses between $I_q$ and each retrieved RGBD image for global localization.


\subsection{Image Retrieval from the Map}
We take advantage of the success in visual place recognition technologies \cite{torii201524,arandjelovic2016netvlad, tang2020adversarial} and utilize NetVLAD \cite{arandjelovic2016netvlad} to extract the global image descriptor for each map image offline. During localization, we extract the global image descriptor for the query image $I_q$ and find the nearest $N$ map images $\{I_{r_i}|i=1,\dots,N\}$ according to the Euclidean distance between image descriptors. As the global pose $T^M_{r_i}$ is known for each retrieved map image $I_{r_i}$, the global pose of $I_q$ can be calculated based on the estimated relative transformation $T^{q}_{r_i}$ between $I_{r_i}$ and $I_q$:
\begin{equation}
T^M_{q_i}=T^M_{r_i} \cdot (T^{q}_{r_i})^{-1}
\end{equation}

In the following sections we introduce our regularization based network designed to calculate the relative transformation and a validation method used to select the best regressed result out of the $N$ estimated poses $\{T^M_{q_i}|i=1,\dots,N\}$.

\subsection{Regularization based Relative Pose Regression}
\subsubsection{\textbf{Motivation}}
Many RPR methods utilize the concatenation of two CNN features as input for regression in the following Fully Connection (FC) layer \cite{laskar2017camera,balntas2018relocnet, ding2019camnet}. In this way, the output of the FC layer is related to the absolute values of the feature pair. Imagining that given two images of the same content, if some patches of the images are changed simultaneously, their feature pair would change accordingly while the regressed pose should not. This brings difficulty to network learning as enormous input features correspond to the same output. Furthermore, the values of the image features are scene-specific, thus making the network difficult to localize in unseen environment.
On the other way, the correlation score between two features only depends on their difference, which should stay stable as long as the feature descriptors are consistent. 
Thus explicitly adding the matching layer within RPR networks could largely reduce the complexity of the pose regression problem, and endow the network with better generalization ability.

Traditionally the correlation volume contains the entire matching information between pixel pairs from two images. When the dynamics is dominant or the perspective of view is largely different, the valid overlap between two images is limited and the correlation volume is occupied with a major of confusing information. 
Different from traditional methods that extract pixel-to-pixel correspondences to satisfy geometry based pose solvers \cite{rocco2018neighbourhood,zhou2020learn,rocco2020efficient}, we extract the matching information implicitly by regularizing the correlation volume with a bottleneck structure based CNN model for dimension reduction as shown in Fig. \ref{framework}. Circumventing pixel-level correspondence for pose estimation brings us with two benefits: i) No pixel-level supervision is required thus the training data is easier to obtain. ii) There is no need to find correspondences at high resolution for pose estimation solved by geometry based methods, thus effective global correlation is accessible due to restricted image resolution with large receptive field, leading to more robust patch-wise matching.

\subsubsection{\textbf{Network Architecture}}
The details of the regularization based RPR network is demonstrated in Fig. \ref{framework}. 
We utilize the pre-trained VGG16 \cite{simonyan2014very} network for feature extraction and truncate it at the last pooling layer \cite{melekhov2019dgc,truong2020glu}. The input image is resized to $256\times256$ before put into network. We only use the last two layers of the features computed by VGG16 with resolution 16$\times$16 (feature $\bm{f}_1\in\mathbb{R}^{16\times16\times C_1}$) and 32$\times$32 (feature $\bm{f}_2\in\mathbb{R}^{32\times32\times C_2}$) for the following two-stage pose estimation. 

In the first layer $\mathcal{L}_1$, the global correlation $\bm{c}^1\in \mathbb{R}^{16\times16\times16\times16}$ between features $\bm{f}^{r_i}_1$ and $\bm{f}^q_1$ of $I_{r_i}$ and $I_q$ respectively is first computed according to the scalar product between each pixel $\bm{u}\in\mathbb{Z}^2$ of $I_{r_i}$ and the corresponding pixel $\bm{u}'$ of $I_q$
\begin{equation}
\bm{c}^1(\bm{u},\bm{u}')={\bm{f}^{r_i}_1(\bm{u})}^T\bm{f}^{q}_1(\bm{u}')
\end{equation}
which then is forwarded into the NC matching layer \cite{rocco2018neighbourhood} to constrain geometric consistency. The following MotionNet module takes the output score map to regress the initial relative pose $T^{q_1}_{r_i}$. Different from other works that represent the pose with 3D translation and a 3D/4D rotation vector \cite{laskar2017camera, ding2019camnet}, 
we represent the pose with a 9D vector $\bm{\xi} = [\bm{r}\in \mathbb{R}^6, \bm{t}\in \mathbb{R}^3]$, where $\bm{t}$ denotes the translation and $\bm{r}$ denotes the rotation \cite{zhou2019continuity}. A mapping $\Phi$ is adapted to transform our rotation vector to conventional rotation matrix, 
\begin{equation}
\Phi:\mathbb{R}^6\rightarrow \mathbb{SO}(3), \bm{r}\mapsto R = \Phi(\bm{r})
\end{equation}
Combined with $t$, $\Phi$ induces a mapping $\tilde{\Phi}$ from our 9D vector to the standard Euclidean transformation
\begin{equation}
\tilde{\Phi}: \mathbb{R}^6\times \mathbb{R}^3 \rightarrow \mathbb{SE}(3), (\bm{r}, \bm{t})\mapsto T = \tilde{\Phi}(\bm{r}, \bm{t})
\end{equation}

We utilize depth image $D_{r_i}$ to calculate rigid flow $\bm{w}$ between $I_q$ and $I_{r_i}$ at the same resolution as $\bm{f}_2$ according to $T^{q_1}_{r_i}$  
\begin{equation}
	\begin{split}
\left[\begin{matrix}
\bm{p}^{q_1}_k\\
1
\end{matrix}\right]&=T^{q_1}_{r_i}\left[\begin{matrix}
K^{-1} \left[\begin{matrix}
\bm{u}_k\\
1 
\end{matrix}\right]  \cdot z_k\\
1
\end{matrix}\right]\\
\bm{w}_k&=\pi(\bm{p}^{q_1}_k)-\bm{u}_k
\end{split}
\end{equation}
in which $ T^{q_1}_{r_i}=\tilde{\Phi}(\bm{\xi^{q_1}_{r_i}}) $, $\bm{u}_k$ denote the $k$th pixel on $I_{r_i}$, $z_k$ and $\bm{p}^{q_1}_k$ are the corresponding depth value and 3D point transformed by $T^{q_1}_{r_i}$. $K$ denotes the intrinsic matrix and $\pi(\cdot)$ denotes the projection function. 

In the second layer $\mathcal{L}_2$, we utilize $\bm{w}$ to warp $\bm{f}^q_2$ and compute its correlation with $\bm{f}^{r_i}_2$. In this layer only local correlation $\bm{c}^2\in\mathbb{R}^{32\times32\times(2n+1)}$ within $n$ neighborhood pixels are searched to refine the matching information calculated on $\mathcal{L}_1$. And this correlation results are forwarded to regress relative pose $\delta T=\tilde{\Phi}(\bm{\xi}) $ as refinement. The final estimated pose in $\mathcal{L}_2$ is
\begin{equation}
 T^{q_2}_{r_i}=T^{q_1}_{r_i}  \delta T
\end{equation}

The only supervision during training is the groundtruth relative pose $T^{q_{gt}}_{r_i}\in \mathbb{SE}(3)$ between $I_q$ and $I_{r_i}$. The pose estimation error in each layer $\mathcal{L}_l (l=1,2)$ is calculated as
\begin{equation}
\Delta T_l = \left[
\begin{matrix}
\Delta R_l & \Delta t_l\\
\mathbf{0_{1\times 3}} & {1} 
\end{matrix} \right]=T^{q_l}_{r_i} {T^{q_{gt}}_{r_i}}^{-1}
\end{equation}
We convert the rotational error into the angular value $\Delta\theta$ and the total loss is defined as 
\begin{equation}
Loss = Loss_1+\beta \cdot Loss_2
\end{equation}
where
\begin{equation}
\begin{split}
\Delta \theta_l &= arccos(\frac{tr(\Delta R_l)-1}{2}) \\
Loss_l &=  \|\Delta \theta_l\|+ \gamma_l \|\Delta \mathbf{t}_l\|
\end{split}
\end{equation}
in which $\beta$ and $\gamma_l$ represent the corresponding weights.




\subsubsection{\textbf{Detail of MotionNet}} In this module the correlation information is regularized for pose regression.
It takes the correlation volume from global or local correlation modules as input and utilize a CNN network with bottleneck structure along the feature dimension and a FC layer to regress the corresponding relative poses as shown in Fig. \ref{framework}. The feature dimension of the score map is regularized into a compact formation, on which the depth image with the same resolution is concatenated for scale recovery.


\subsection{Correlation based Pose Selection}
After calculating the relative poses between $I_q$ and the $N$ retrieved map images, we evaluate the $N$ results according to the correlation between $\bm{f}^{r_i}_1$ and warped $\bm{f}^q_1$ based on the rigid flow computed by $T^{q_2}_{r_i}$. We apply softmax along the channels of the correlation results and count the number of vectors in which the highest correlation score is larger than threshold $\alpha$. Only valid correlations that the warped positions are within the images are counted. The pose of the image pair with max number is selected as the best regressed result.






\section{Experiments }

In this section, we assess the performance of our proposed regularization based RPR framework \footnote{https://github.com/syywh/RRPR} with standard 7Scenes \cite{shotton2013scene} dataset for comparison with other networks, and also utilize a challenging indoor public dataset OpenLoris-Scene \cite{shi2020we} to investigate the potential of regression based methods addressing complex real-world environments compared with the geometry based MPL method \cite{jiao20202}. 

\subsection{Datasets and implementation details}
\textbf{OpenLoris-Scene} \cite{shi2020we} is a public indoor dataset designed to evaluate the performance of lifelong visual SLAM methods. 
It collects RGBD image sequences using a mobile robot in five scenes separately along various trajectories and situations. The data includes significant appearance, illumination and perspective of view changes as well as textureless areas and blur, which is valuable to access the performance of vision based methods facing real-world situation.
We utilize the first three sequences in both ``Home'' and ``Office'' scenes for training. Any two images from same or different sequences are selected as a training image pair if their translation and orientation distances are within the set thresholds. The translational threshold is set to $1.5m$ and rotational threshold is $30^{\circ}$. We test the localization performance on the other sequences in ``Home'' and ``Office'' scenes. To evaluate the performance of generalization, we also utilize the sequences in the ``Cafe'' scene for testing, in which the environmental appearance is entirely unseen in the training data.

\textbf{7Scenes} \cite{shotton2013scene} contains RGBD images collected from 7 indoor rooms. We utilize the training data listed in 7 scenes together for training and compare our testing results with the other learning based pose estimation methods. We also evaluate the cross-dataset generalization performance with 7Scenes dataset for comparison. 

\textbf{TUM-RGBD} \cite{sturm2012benchmark} dataset contains sequential RGBD images collected in different scenes. We want to evaluate cross-dataset generalization based on the models trained OpenLoris-Scene and 7Scenes datasets, but the training data in OpenLoris-Scene only contains planar motion. We finetune the models on TUM-RGBD dataset to supplement the freedom of motion in training data of OpenLoris-Scene.

\textbf{Implementation details: }The network is implemented in PyTorch \cite{paszke2017automatic} and trained for at most 50 epochs with the weights of VGG16 feature extraction layer fixed. We would stop the training process early if the training loss does not decrease. All the models are trained using AdamOptimizer \cite{kingma2014adam} with beginning learning rate of $1e^{-4}$ and decay ratio of $0.7$ every 10 epochs. 
The batch size is set to 6 and $\beta=4, \gamma_1=3, \gamma_2=2, n=4, \alpha=0.007$. 

\textbf{Network structure details:} 
To validate the effectiveness of the matching layer and correlation regularization process, we implement three types of RPR networks within our proposed two-layer framework with difference in the input feature for pose regression: 
\begin{enumerate}
\item image feature concatenation (``feature-cat'' in tables)
\item correlation volume output from the NC matching layer (``score-map'' in tables)
\item correlation volume with dimension regularization (``score-map-dr$\bm{x}$'' in tables, $\bm{x}$ denotes the channel of the compact regularized feature)
\end{enumerate}
For fair comparison all the three networks have CNN modules between the input features and the pooling  layer within each MotionNet.

\subsection{Localization performance}

\begin{table*}[htbp]
	\setlength{\itemsep}{0pt}
	\setlength{\parskip}{0pt}
	\centering
	\caption{Visual Localization results evaluated on 7Scenes datasets.}
	\setlength{\tabcolsep}{0.6mm}{
		\begin{threeparttable}
		\begin{tabular}{ccccccccccccc}
			\toprule
			& \multicolumn{1}{c|}{MPL} & \multicolumn{3}{c|}{APR}  & \multicolumn{8}{c}{RPR}  \\
			& \multicolumn{1}{c|}{sift+5p}& PoseNet & MapNet &  \multicolumn{1}{c|}{DSAC++} & RelativePN& RelocNet & \multicolumn{1}{c}{NC-EssNet}& {score-map}$\   $ &  {score-map} & {score-map-dr4} & \multirow{2}{*}{score-map} & \multirow{2}{*}{feature-cat} \\
			& \multicolumn{1}{c|}{\cite{zhou2020learn}}& \cite{kendall2015posenet} & \cite{brahmbhatt2018geometry} & \multicolumn{1}{c|}{ \cite{brachmann2018learning}} & \cite{laskar2017camera} & \cite{balntas2018relocnet} &  \multicolumn{1}{c}{\cite{zhou2020learn}} &-dr2 &-dr3 &-dr4 &  &  \\
			\midrule
			median error & \multirow{2}{*}{1.99/0.08} & \multirow{2}{*}{10.44/0.44} & \multirow{2}{*}{7.78/0.21} & \multirow{2}{*}{1.10/0.04} & \multirow{2}{*}{9.30/0.21} & \multirow{2}{*}{6.74/0.21} & \multirow{2}{*}{7.50/0.21} & L1: 3.85/0.14 & 3.79/0.15&3.99/0.15 & 4.34/0.16 & 9.55/0.37   \\
			(deg/m) &   &    &       &       &       &       &       &L2: 3.22/0.11  & 3.37/0.12 & 3.26/0.11 & 3.35/0.12 & 8.85/0.36  \\
			\bottomrule
					\end{tabular}%
			\begin{tablenotes}
				\item  ``L1'' denotes the outputs from the first layer $\mathcal{L}_1$ and ``L2'' denotes the outputs from layer $\mathcal{L}_2$ in our framework
			\end{tablenotes}
	\end{threeparttable}
	}
	\label{tab:7Scenes}%
\end{table*}%

\begin{table*}[htbp]
	\centering
	\caption{Visual localization results evaluated on OpenLoris-Scenes datasets. Each column lists the percentage of localization results satisfying the translational and rotational error thresholds ($0.25m,5\deg$)/($0.5m,5\deg$)/($1m,5\deg$). }
	\begin{threeparttable}
	\begin{tabular}{cccccccc}
		\toprule
		& Home1-4 & Home1-5 & Office1-4 & Office1-5 & Office1-6 & Office1-7 & Cafe2-1 \\
		\midrule
		\midrule
		score-map-dr2 (gt select) &53.8/61.0/63.7  &90.2/\textcolor{blue}{92.4/92.8}  &12.0/20.1/24.7  & 29.3/31.1/33.0 & 91.4/96.5/96.8 &0/0.1/5.7  &72.1/78.7/\textcolor{blue}{83.0}  \\
		score-map-dr3 (gt select) &53.2/59.6/62.8 &89.7/91.3/91.5 &13.8/23.4/{34.5} &29.5/31.7/35.8 &89.6/98.5/98.5 &\textcolor{blue}{9.1}/19.1/28.7 &70.7/76.2/82.2 \\
		score-map-dr4 (gt select) &57.3/60.9/62.7 & 90.0/91.5/91.5 &{18.3/29.7}/32.4 &{30.5/34.9/37.4} &91.4/96.8/97.2 &3.2/\textcolor{blue}{20.7/30.6} & \textcolor{blue}{74.5/81.8}/82.2 \\
		score-map (gt select) & \textcolor{blue}{58.0/63.6/64.4} & \textcolor{blue}{91.8}/92.3/92.3 & 2.1/7.4/13.9 & 27.6/28.9/30.5 & 90.5/99.4/99.6 & 0/0/0.1 & 72.2/80.4/82.4 \\
		feature-cat (gt select) & 48.7/55.7/62.1 & 78.7/84.2/85.5 & \textcolor{blue}{26.4/32.6/36.3} & 26.4/32.4/37.7 & 84.4/97.5/98.6 & 0/0.1/11.3  & 57.6/64.5/67.4 \\
	\midrule
			score-map-dr2 (corr select) &43.1/49.9/51.6 &82.7/85.8/85.9 & 7.5/15.2/18.0 &26.2/28.4/30.0 &88.2/91.6/91.6 &0/0/4.7 &65.6/76.1/82.3 \\
		score-map-dr3 (corr select) &43.4/49.6/52.2 &84.1/85.6/85.6 &10.1/18.7/22.1 &28.0/28.8/30.7 &87.3/93.9/94.0 &\textcolor{red}{5.2/11.7/20.2} &66.9/73.4/81.3 \\
		score-map-dr4 (corr select) &47.3/52.9/53.5 &85.8/86.8/86.8 &10.5/\textcolor{red}{21.1/24.0} &27.9/29.3/30.7 &89.4/95.0/95.0 &1.6/7.7/23.0 &\textcolor{red}{71.3/80.4}/80.6 \\
        score-map (corr select) &\textcolor{red}{50.1/57.2/58.1} &\textcolor{red}{86.6/86.9/86.9} &1.1/4.4/8.3 &27.3/27.7/28.4 &87.0/91.3/91.3 &0/0/0 &69.4/78.0/81.9 \\
		feature-cat (corr select) &40.5/49.7/55.5 &69.7/74.1/74.1 &{13.6}/20.0/22.4 &23.7/28.4/32.9 &73.4/90.5/91.4 &0/0/6.7 & 52.9/61.4/62.1 \\
		\midrule
		SuperPoint+2p \cite{jiao20202} & 38.3/48.5/53.6 & 84.2/86.0/86.0 & \textcolor{red}{15.3}/20.9/20.9 & \textcolor{red}{39.2/39.2/39.3} & \textcolor{red}{100/100/100} & 0.1/0.5/0.5 & 55.3/79.6/\textcolor{red}{85.9} \\
		\bottomrule
	\end{tabular}%
\begin{tablenotes}
	\item The best results are marked as red, and the best results selected by groundtruth are marked as blue.
\end{tablenotes}
\end{threeparttable}
	\label{tab:OpenLoris}%
\end{table*}%

\begin{table*}[htbp]
	\centering
	\caption{Generalization results of different RPR methods evaluated on 7Scenes.}
	\begin{tabular}{cccccccc}
		\toprule
		method& score-map-dr2 & score-map-dr3 & score-map-dr4 & {score-map} & {feature-cat} & RelativePN \cite{laskar2017camera} & {RelocNet}\cite{balntas2018relocnet} \\
		training dataset  & TUM\cite{sturm2012benchmark}   &TUM\cite{sturm2012benchmark} &TUM\cite{sturm2012benchmark}   & TUM\cite{sturm2012benchmark}   & TUM\cite{sturm2012benchmark}   & University\cite{laskar2017camera} & ScanNet\cite{dai2017scannet} \\
			\midrule
		median error(deg/m) & 6.37/0.24 &5.84/0.23  &5.89/0.23 & 5.71/0.21 & 10.5/0.30 & 18.37/0.36 & 11.29/0.29 \\
		\bottomrule
	\end{tabular}%
	\label{tab:generalization}%
\end{table*}%

We first compare our methods with state-of-the-art methods in terms of localization accuracy by training and testing the models on OpenLoris-Scenes and 7Scenes datasets separately. As there is no visual localization benchmark on OpenLoris-Scenes datasets, we implement 2p-RANSAC based MPL solution \cite{jiao20202} with SuperPoint features \cite{detone2018superpoint} as the baseline. Note that as there is only planar motion in OpenLoris-Scenes datasets, 2p-RANSAC based solution should outperform traditional EPnP \cite{lepetit2009epnp} or 5p \cite{nister2004efficient} based RANSAC solvers. We set the first sequences in ``Home'' and ``Office'' as the maps for each scene and for images in the query sequences we utilize NetVLAD \cite{arandjelovic2016netvlad} to retrieve top-5 images as the reference images. For 2p-RANSAC based solution we merge the features in all of the retrieved 5 images to do pose estimation for better robustness. While for our RPR methods we regress the relative pose with each retrieved image separately and select out one result according to the evaluation method in III-C (``corr select'' in TABLE \ref{tab:OpenLoris}). We also list the results selected by the groudtruth (``gt select'' in TABLE \ref{tab:OpenLoris}) that chooses the result with smallest localization error, which can be considered as the best performance the RPR networks achieve. While for 7Scenes datasets, we retrieve the top-1 image for RPR evaluation and compare the results with the other MPL \cite{zhou2020learn} and learning based localization methods \cite{zhou2020learn,kendall2015posenet,laskar2017camera,brahmbhatt2018geometry,balntas2018relocnet,brachmann2018learning}. The localization results are shown in TABLE \ref{tab:7Scenes}, \ref{tab:OpenLoris}.

From the evaluation results on TABLE \ref{tab:7Scenes}, we can see that our proposed models implemented with matching layers (``score-map-dr2, score-map-dr3, score-map-dr4, score-map'') outperform the other listed RPR based methods and part of APR based methods. Note that we also outperform the result in \cite{zhou2020learn} which also leverages matching layer for pose regression and we owe it to the pyramid structure based image scale regularization and the combination with depth information. As there is tiny appearance change or dynamics within the environments, MPL and scene coordinate based methods \cite{brachmann2018learning} could achieve superior localization precision as adequate pixel-level correspondences could be found, but our methods also show comparable performance. In TABLE \ref{tab:7Scenes} we list the outputs from the two layers in our frameworks to demonstrate the effectiveness of the second layer in terms of precision improvement. In other experiments we only list the output of the second layer as the regression results.

\begin{figure}
	\centering
	\includegraphics[width=0.47\textwidth]{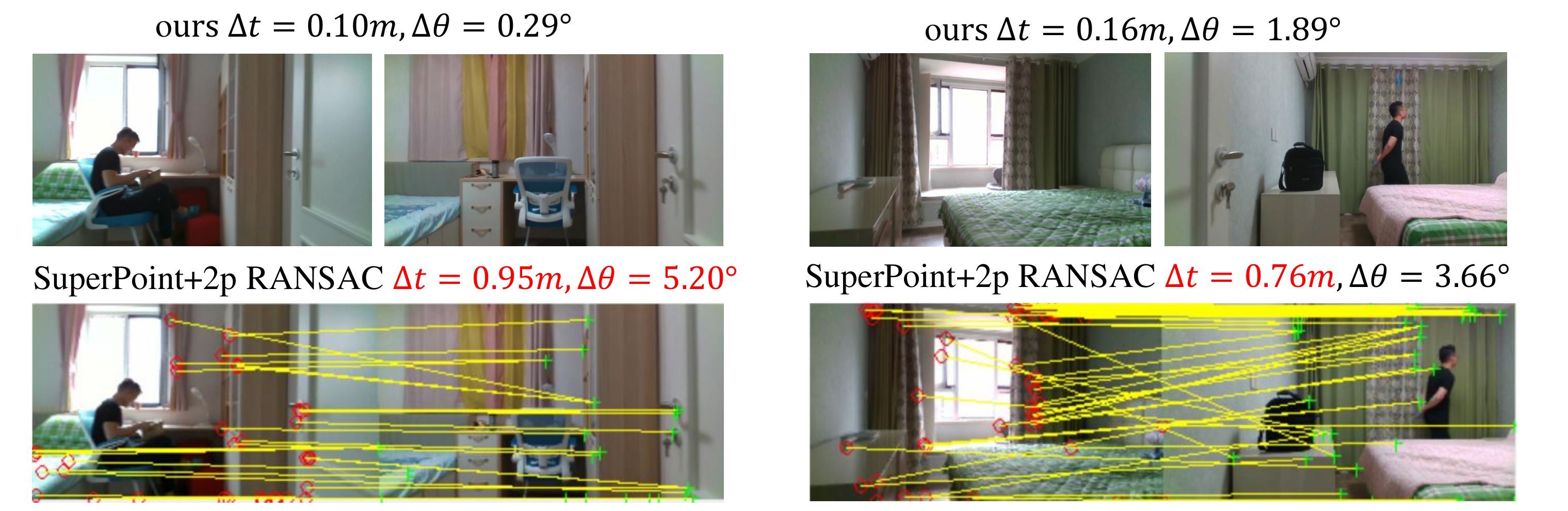}
	\caption{Two cases that SuperPoint+2p-RANSAC method fails but our proposed RPR methods success in pose estimation. In second row we draw the match results computed by SuperPoint descriptors with yellow lines.}
	\label{bettercase}
\end{figure}

\begin{figure}
	\centering
				\includegraphics[width=0.4\textwidth]{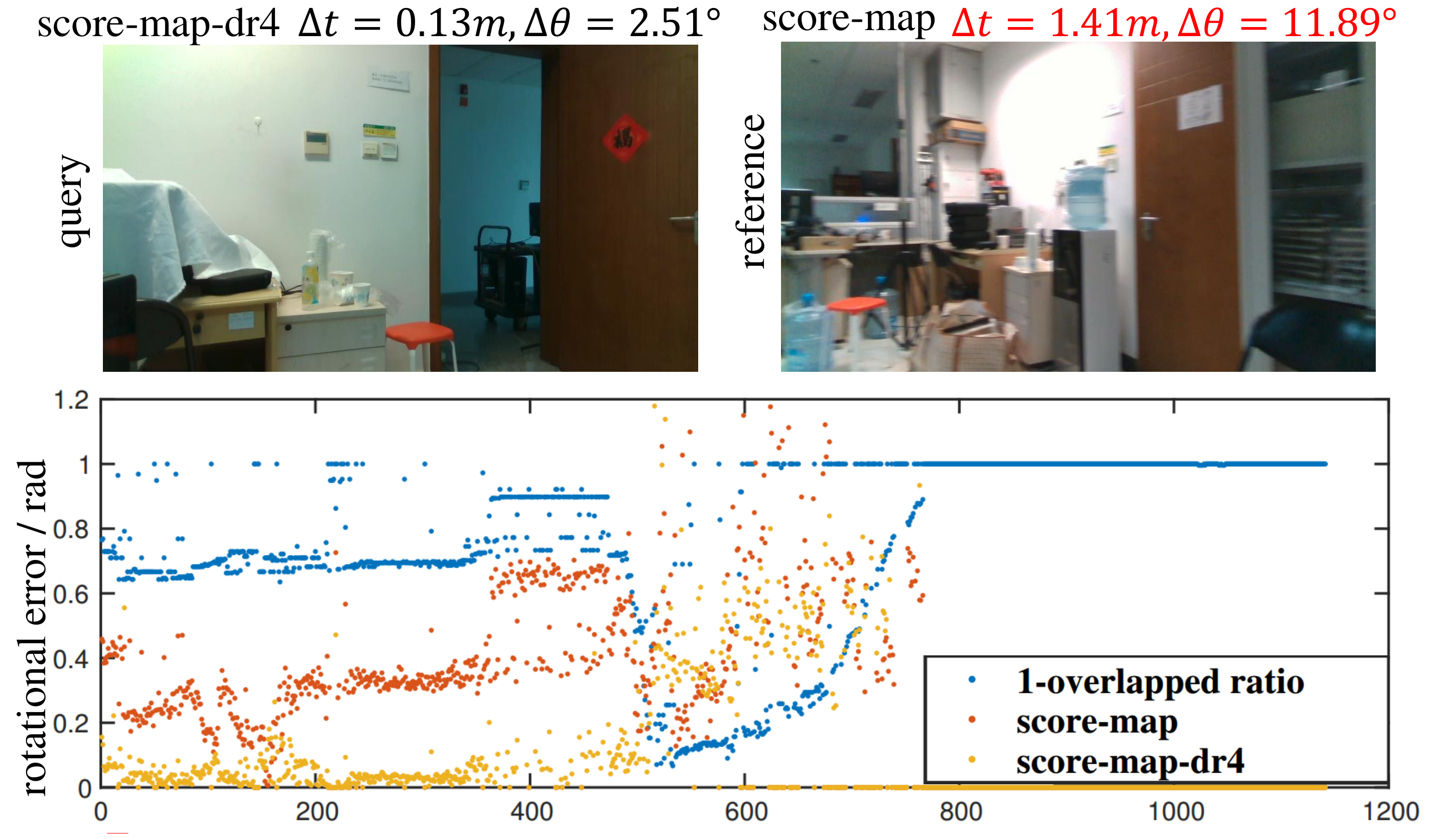}
			\caption{Rotational errors of different methods evaluated in ``Office1-7''. For better visualization we set the results of the image pairs with overlapped ratio less than 0.1 as 0. The two images are related to the data indicated by the green arrow.}
	\label{office-error}
\end{figure}

\begin{figure}
	\centering
	\includegraphics[width=0.47\textwidth]{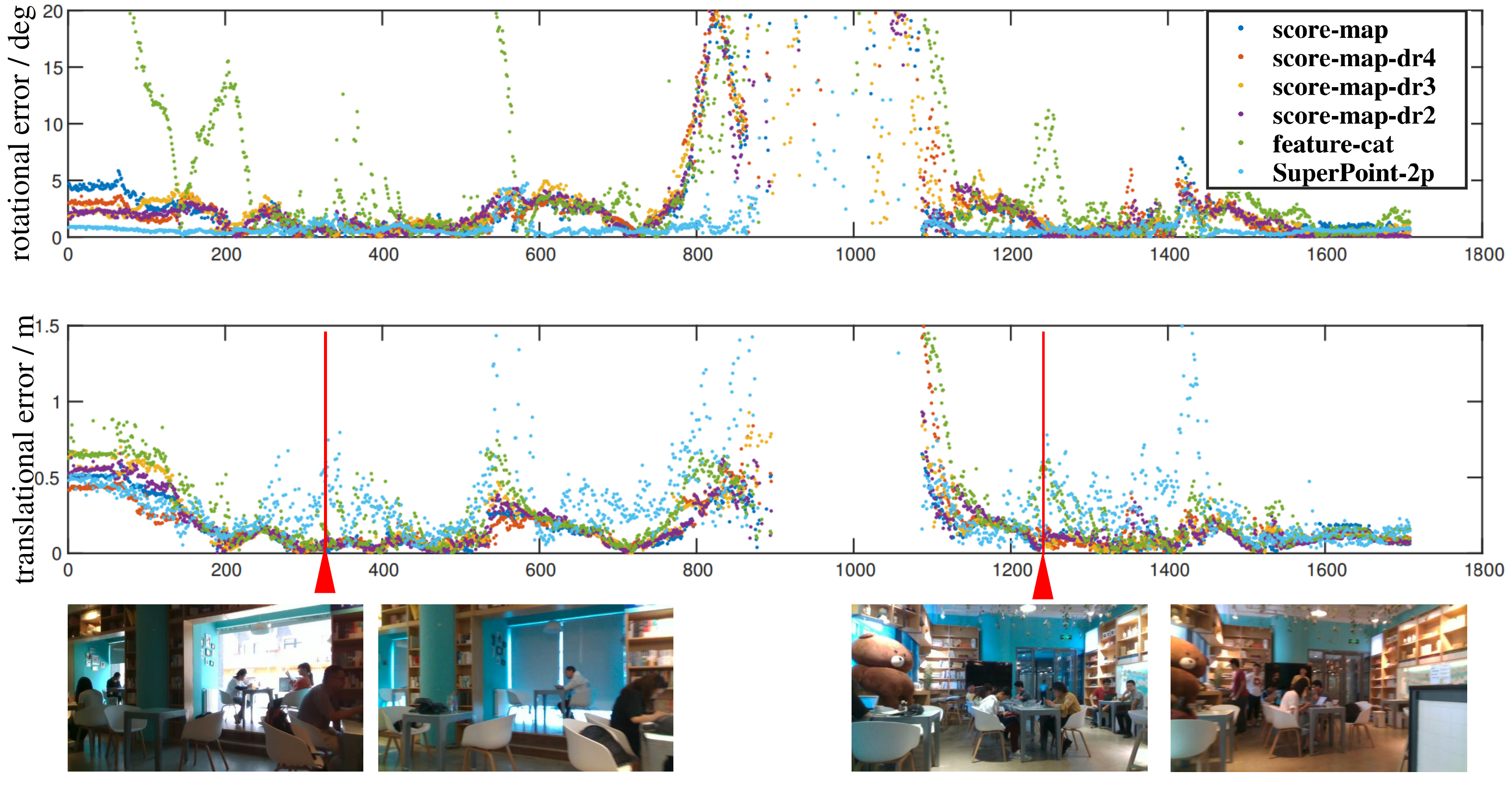}
\caption{The generalization results tested in ``Cafe2-1'' with different methods. The image pairs related to the red arrows are listed in the second line as reference to indicating the environment.}
	\label{cafe-error}
\end{figure}

TABLE \ref{tab:OpenLoris} includes the localization results on OpenLoris-Scenes dataset. As we only use one sequence for each scene as the map, some query images cannot find the matched reference images thus for some sequences the best performance cannot achieve 100$\%$. We can see that in most of the localization sequences our RPR based methods could outperform SuperPoint+2p-RANSAC method. 
We analyze through the failure cases in scenes with large different performance to find the strengths and weaknesses of each methods. In ``Home1-4'' scene where RPR based methods extremely outperform RANSAC based method, we find that many failure cases occur in places that the appearance changes significantly or many dynamics exist.   
We draw two cases that SuperPoint+2p-RANSAC method fails but RPR methods could give good localization results in Fig. \ref{bettercase}. We can see that many matches are inaccurate, and some matched points belong to dynamic objects such as the curtain, which would further degrade the pose estimation results. While in the proposed RPR methods, as the utilized image features have large receptive field and global correlation information is leveraged, accurate pose estimation could be achieved even though there are many dynamics in the views.    

In ``Office1-4'' and ``Office1-7'', we find that RPR method without correlation regularization shows largely decreased performance compared with the results with correlation regularization. In this two cases, the trajectories of query sequences are almost opposite to the mapping trajectories, and some objects are removed across scenes. Thus many retrieved images only have little overlap with the query ones. We draw the rotational errors of ``score-map'' and ``score-map-dr4'' tested in ``Office1-7'' along with the overlap ratio  in Fig. \ref{office-error}. We can see at the beginning of the trajectory in which the environments can be inferred from figures in the first row of Fig. \ref{office-error}, ``score-map-dr4'' shows exceeding performance facing large perspective of view, which validates the effectiveness of our proposed correlation reduction process.

\subsection{Generalization Study}

\begin{table}[]
\centering
\caption{Generalization results evaluated on OpenLoris datasets based on the models trained on 7Scenes. Each column lists the percentage of localization results satisfying the translational and rotational error thresholds ($0.25m,5\deg$)/($0.5m,5\deg$)/($1m,5\deg$). }
\begin{threeparttable}
\begin{tabular}{lcc}
	\toprule
	& Office1-6 & Cafe2-1 \\
	\midrule
			score-map-dr2 (7S, gt select)  &\textcolor{blue}{87.7/90.5/90.8} &64.6/74.6/77.2 \\
			score-map-dr3 (7S, gt select) & 78.8/84.3/85.1& \textcolor{blue}{71.1/82.4/83.5}\\
			score-map-dr4 (7S, gt select) &78.8/84.3/85.1 &65.5/70.3/71.0 \\
			score-map (7S, gt select) &74.2/81.5/81.5 &{67.7}/72.5/73.2\\
			feature-cat(7S, gt select)  &55.8/75.3/77.3 &22.2/41.0/47.2 \\
	\midrule
			score-map-dr2 (7S, corr select)  &\textcolor{red}{78.9/79.9/79.9} &55.9/64.5/65.3 \\
			score-map-dr3 (7S, corr select) &64.0/67.9/67.9  &58.0/\textcolor{red}{76.2/77.3} \\
			score-map-dr4 (7S, corr select) &64.0/67.9/67.9  &58.1/62.4/62.7 \\
			score-map (7S, corr select) &52.9/60.5/60.5 &  \textcolor{red}{59.0}/66.0/65.0\\
			feature-cat(7S, corr select) &34.8/50.6/50.9 & 11.1/24.9/27.9\\
	\midrule
	SuperPoint+2p \cite{jiao20202}  & {100/100/100} & 55.3/ {79.6/85.9} \\
	\bottomrule
\end{tabular}%
\begin{tablenotes}
	\footnotesize
\item The best results of RPR methods are marked as red and the best results selected by groundtruth are marked as blue.
\end{tablenotes}
\end{threeparttable}
\label{tab:OpenLoris-general}%
\end{table}%

In this subsection we execute generalization study of our networks and evaluate the performance of different input feature structures for the regression layer. We reuse the trained models on ``Home'' and ``Office'' scenes to test the cross-scene generalization performance on the ``Cafe'' scene, and the results are listed in the last column of TABLE \ref{tab:OpenLoris}. Then we finetune these models on TUM-RGBD dataset for 5 epochs to supplement the degree of motion before applying them on 7Scenes dataset to test the cross-dataset generalization performance. Our results as well as the other generalization results from the state-of-the-art methods are shown in TABLE \ref{tab:generalization}.   

From the generalization results in TABLE \ref{tab:OpenLoris}, we find that our RPR networks with matching layer still outperform SuperPoint+2p-RANSAC method at the first two error intervals, which validates the generalization ability of our RPR networks. In this case the result of ``feature-cat'' network largely degrades, which reflects the importance of the matching layer in terms of improving generalization performance. We draw the localization errors of different methods in Fig. \ref{cafe-error} and the results show that most of the failure cases of SuperPoint+2p-RANSAC method are due to large translational error. We select two places from these failure cases and show the image pairs on Fig. \ref{cafe-error}. We can find that in these cases though the appearance change is not as significant as in ``Home'', there are many dynamic objects and people in the scene, making it hard to find accurate pixel-level correspondences with reliable depth for pose estimation, which also reveals the advantage of regression based methods that release the requirement for pixel-to-pixel correspondence.

TABLE \ref{tab:generalization} shows the generalization performances of different methods tested on 7Scenes and we can find that the precision of our proposed methods with matching layer even outperform the results of some RPR methods listed in TABLE \ref{tab:7Scenes}. We also test the generalization performance of the models trained on 7Scenes to OpenLoris-Scenes. As there is no appearance change in the 7Scenes datasets, we only show the generalization results on ``Office1-6'' and ``Cafe2-1'' in which the query and mapping trajectories are almost the same and only small part of the scenes are changed. The results are listed in TABLE \ref{tab:OpenLoris-general}. The experimental results show that even the sensors and environments are both changed, the degeneration problem is not severe.

\subsection{Ablation Study}
Here we evaluate the effectiveness of depth fusion in terms of scale recovery. 
To validate the effectiveness of scale recovery of our method, we train the network with no depth concatenation on the regularized feature and test its generalization performance on ``Cafe'' dataset. Besides listing the results according to the thresholds in TABLE \ref{tab:OpenLoris}, we also list the percentage of the results that with angular error smaller than $5^{\circ}$ in TABLE \ref{tab:depth_ablation}. From the results we can find that the performances of rotational estimation are similar between methods with and without depth concatenation, while the performances taking translational error into account differ a lot, which can demonstrate that the depth concatenation within the model can successfully contribute to recover scale.

\begin{table}[]
	\centering
	\caption{Ablation results to validate the effectiveness of scale recovery tested on Cafe2-1 dataset.}
	\setlength{\tabcolsep}{0.6mm}{
		\begin{tabular}{ccccc}
			\toprule
			& score-map-dr2 & score-map & feature-cat & dr2-noDepth \\
			\midrule
			median error    & 72.1/78.7/83.0 & 72.2/80.4/82.4 & 57.6/64.5/67.4 & 56.4/63.9/66.6 \\
			$\Delta\theta_2<5^{\circ}$  & 84.1  & 83.8  & 68.3  & 83.8 \\
			\bottomrule
		\end{tabular}%
	}
	
	\label{tab:depth_ablation}%
\end{table}%


\section{CONCLUSIONS}
In this paper we propose a novel relative pose regression framework for visual localization. In order to improve the network generalization towards unseen scenes, we explicitly add a matching layer and utilize the correlation volume for pose regression. Besides, we design a pyramid based structure to regress the pose from coarse to fine with restricted resolution, and apply dimension reduction on the correlation channel to improve the robustness facing large perspective of view. The experiments validate that our network could achieve state-of-the-art localization performance and demonstrate comparable results even in generalization test to unseen scenes. Experiments also show that regression based visual localization methods possess large potential in complicated real-world environments compared with the methods that require pixel-level correspondences for pose estimation. In the future we design to do research on fusing multiple map images to improve the robustness of visual localization.

\addtolength{\textheight}{-3cm}   




\bibliographystyle{ieeetr}
\bibliography{library}

\end{document}